\documentclass{article}

\usepackage{microtype}
\usepackage{graphicx}
\usepackage{subfigure}
\usepackage{booktabs} 
\usepackage{multicol,multirow}
\usepackage{bbding}
\usepackage{amsmath}
\usepackage{amssymb}
\usepackage{mathtools}
\usepackage{amsthm}
\usepackage{tcolorbox}
\definecolor{dkblue}{rgb}{0,0.08,0.45}

\theoremstyle{plain}
\newtheorem{theorem}{Theorem}[section]
\newtheorem{proposition}[theorem]{Proposition}

\theoremstyle{definition}

\newtheorem{assumption}[theorem]{Assumption}

\theoremstyle{remark}

\DeclareMathOperator*{\argmin}{arg\,min}
\DeclareMathOperator*{\argmax}{arg\,max}
\DeclareMathOperator*{\bias}{\mathrm{Bias}}

\newcommand{\meanN}{\frac{1}{n} \sum_{i=1}^n}

\newcommand{\sumA}{\sum_{a \in \mathcal{A}}}
\newcommand{\sumE}{\sum_{e \in \mathcal{E}}}

\newcommand{\mE}{\mathbb{E}}
\newcommand{\mV}{\mathbb{V}}

\newcommand{\mR}{\mathbb{R}}

\newcommand{\ind}{\mathbb{I}}
\newcommand{\calD}{\mathcal{D}}
\newcommand{\calX}{\mathcal{X}}
\newcommand{\calA}{\mathcal{A}}
\newcommand{\calE}{\mathcal{E}}

\newcommand{\calQ}{\mathcal{Q}}
\newcommand{\calN}{\mathcal{N}}
\newcommand{\indep}{\perp \!\!\! \perp}

\newcommand{\mips}{\hat{V}_{\mathrm{MIPS}} (\pi_e; \calD)}
\newcommand{\mdr}{\hat{V}_{\mathrm{MDR}} (\pi_e; \calD, \hat{q})}

\usepackage{hyperref}

\usepackage[accepted]{icml2023}

\icmltitlerunning{Off-Policy Evaluation and Learning under Time-Series Non-Stationarity}

\begin{document}

\twocolumn[
\icmltitle{Doubly Robust Estimator for Off-Policy Evaluation with Large Action Spaces}




\begin{icmlauthorlist}
\icmlauthor{Tatsuhiro Shimizu}{waseda}
\icmlauthor{Laura Forastiere}{yale}
\end{icmlauthorlist}

\icmlaffiliation{yale}{Department of Biostatistics, Yale University, New Haven, CT, United States}
\icmlaffiliation{waseda}{School of Political Science and Economics, Waseda University, Tokyo, Japan}

\icmlkeywords{off-policy evaluation, machine learning, contextual bandit}

\vskip 0.3in
]
\icmlcorrespondingauthor{Tatsuhiro Shimizu}{t.shimizu432@akane.waseda.jp}
\icmlcorrespondingauthor{Laura Forastiere}{laura.forastiere@yale.edu}



\printAffiliationsAndNotice{}  

\begin{abstract}
We study Off-Policy Evaluation (OPE) in contextual bandit settings with large action spaces. The benchmark estimators suffer from severe bias and variance tradeoffs. Parametric approaches suffer from bias due to difficulty specifying the correct model, whereas ones with importance weight suffer from variance. To overcome these limitations, Marginalized Inverse Propensity Scoring (MIPS) was proposed to mitigate the estimator's variance via embeddings of an action. Nevertheless, MIPS is unbiased under the no direct effect, which assumes that the action embedding completely mediates the effect of action on reward. To overcome the dependency on this unrealistic assumptions, we propose a \textit{Marginalized Doubly Robust (MDR)} estimator. Theoretical analysis shows that the proposed estimator is unbiased under weaker assumptions than MIPS while reducing the variance against MIPS. The empirical experiment verifies the supremacy of MDR against existing estimators with large action spaces.
\end{abstract}

\section{Introduction}
Many intelligent systems like recommendation systems \cite{gruson2019offline,narita2020debiased,saito2020unbiased, saito2021counterfactual,joachims2021recommendations} and personalized medicine \cite{gottesman2019guidelines} gradually utilize individual-level data to optimize their decision-making for individuals to enhance the users' experience. One of the ways to attain such an aim is to conduct an A/B test online, but the drawback of such an online algorithm \cite{agarwal2020reinforcement, lattimore2020bandit} is that it is costly and may harm the user experience. Therefore, much research \cite{dudik2014doubly, swaminathan2015counterfactual, wang2017optimal, farajtabar2018more, su2019cab, su2020adaptive, metelli2021subgaussian, saito2022off} has been done to use the past data to evaluate the intervention accurately, which is called \textit{Off-Policy Evaluation (OPE)}. Most of the problems in OPE are formulated in the contextual bandits' settings \cite{wang2017optimal} where we observe context (e.g., the demography of the users), an action collected by the already implemented policy in the system called \textit{behavior policy}, and reward (e.g., the click, purchase of the product, recovery of the patient). In such settings, the goal of OPE is to evaluate the counterfactual performance of the evaluation policy that the system had not implemented. Unfortunately, with large action spaces, which is often the case in the real world, existing estimators suffer from bias due to misspecification of the model or variance due to the wide range of importance weight. To circumvent these limitations, Saito and Joachims \cite{saito2022off}
 proposed Marginalized Inverse Propensity Scoring (MIPS), utilizing importance weight on the space of action embedding to have lower variance. However, the assumption required for the unbiasedness of MIPS does not necessarily hold, as it requires that the embedding of the action fully mediates the effect of the action on the reward, which is not realistic in practice. To alleviate this limitation of MIPS, we develop a doubly robust estimator called MDR, which is unbiased under either the assumption required for MIPS or that on the model while preserving the variance reduction against MIPS. Synthetic data analysis validates that MDR enables us to evaluate a target policy more accurately than the existing estimators.
 
\section{Background}
The data we consider is the contextual vector $x \in \calX \subseteq \mR^{d_x}$, action $a \in \calA$, and reward $r \in [0, r_{\text{max}}]$. We observe the context vector from an unknown distribution $x \sim p(x)$, action $a \sim \pi(a|x)$ from the stochastic intervention called \textit{policy} $\pi: \calX \to \Delta(\calA)$ given contextual vector $x$, and reward $r$ from an unknown distribution $r \sim p(r|x, a)$ given contextual vector $x$ and action $a$. We observe independent and identically distributed $n$ samples collected by the behavior policy $\pi_b$ \cite{strehl2010learning, langford2008exploration}. Thus, the actually observed data called \textit{logged bandit data} $\calD$ is given by
\begin{align*}
    \calD = \{(x_i, a_i, r_i)\}_{i=1}^n \sim \prod_{i = 1}^n p(x_i) \pi_b(a_i|x_i) p(r_i|x_i, a_i).
\end{align*}
If we know the distribution of contextual vector $p(x)$ and reward $p(r|x, a)$, then we can obtain the performance of the policy called \textit{value function} $V(\pi)$ of $\pi$, which means how good the policy $\pi$ is by 
\begin{align*}
    V(\pi):= \mE_{p(x) \pi(a|x) p(r|x, a)}[r] = \mE_{p(x) \pi(a|x)}[q(x, a)]
\end{align*}
where $q(x, a) = \mE_{p(r|x, a)}[r|x, a]$ is the expected reward given contextual vector $x$ and action $a$. However, as we do not know the distribution of contextual vector $p(x)$ and reward $p(r|x, a)$ in practice, we need to estimate the value function to evaluate the performance of a policy $\pi$. We define the \textit{evaluation policy} $\pi_e$ whose value function should be estimated to distinguish it from the behavior policy $\pi_b$. Then, the problem of our interest is how to construct the estimator $\hat{V}(\pi_e; \calD) \approx V(\pi_e)$ where we use the Mean Squared Error (MSE):
\begin{align*}
    \text{MSE}\left(\hat{V}(\pi_e)\right) 
    &= \mE_{\calD}\left[ (V(\pi_e) - \hat{V}(\pi_e; \calD))^2 \right]\\
    &= \text{Bias}(\hat{V}(\pi_e)) + \mV_{\calD}[\hat{V}(\pi_e; \calD)]
\end{align*}
as the quantity to measure how good the estimator is.

\section{Existing Estimators}
There are three classical estimators to estimate the value function $V(\pi_e)$: Direct Method (DM) \cite{beygelzimer2009offset}, Inverse Propensity Score (IPS) \cite{horvitz1952generalization}, and Doubly Robust (DR) \cite{dudik2014doubly}. These estimators have drawbacks when the cardinality of the action space is large. Marginalized Inverse Propensity Scoring (MIPS) plays a significant role in large action spaces to mitigate such shortcomings.
\subsection{Direct Method}
DM \cite{beygelzimer2009offset} uses the estimated expected reward function $\hat{q}: \calX \times \calA \to \mR$ to estimate the value function as follows.
\begin{align*}
    \hat{V}_{\text{DM}}(\pi_e; \calD, \hat{q}) 
    &:= \frac{1}{n} \sum_{i = 1}^n \mE_{\pi_e(a|x_i)} [\hat{q}(x_i, a)] \\
    &= \frac{1}{n} \sum_{i = 1}^n \sum_{a \in \calA} \pi_e(a|x_i) \hat{q}(x_i, a).
\end{align*}
where $\hat{q}(x, a)$ is the estimated expected reward given contextual vector $x$ and $a$. 
For instance, we can consider the following function: $\hat{q} \in \argmin_{q' \in \calQ} \frac{1}{n} \sum_{i = 1}^n \left( 
r_i - q'(x_i, a_i) \right)^2$ where $\calQ$ is some space of the model where we want to optimize $q'$.
 DM is unbiased under the perfect estimation of the expected reward \cite{dudik2014doubly} as follows.
\begin{assumption}[Perfect Estimation of Expected Reward Function]
    \label{unbias_q_x_a}
    We say that the regression model $\hat{q}$ has perfection estimation if $\hat{q}(x, a) = q(x, a)$ for all context $x \in \calX$ and action $a \in \calA$.
\end{assumption}
In practice, it is significantly difficult to accurately estimate the regression model $\hat{q}$, so DM incurs a significant bias in the event of a large action space.

\subsection{Inverse Propensity Scoring}
Unlike the parametric approach like DM, IPS \cite{horvitz1952generalization} re-weights the reward $r_i$ by the ratio $\pi_e(a_i|x_i)/\pi_b(a_i|x_i)$ of the propensity scores of behavior and evaluation policies as follows.
\begin{align*}
    \hat{V}_{\text{IPS}}(\pi_e; \calD) := \frac{1}{n} \sum_{i = 1}^n w(x_i, a_i) r_i
\end{align*}
where $w(x, a)$ is a \textit{vanilla importance weight} defined as $w(x, a) := \pi_e(a|x)/\pi_b(a|x)$.
IPS is unbiased \cite{saito2022off} under the common support defined as follows.
\begin{assumption}[Common Support]
    \label{common_support}
    We say the behavior policy $\pi_b$ satisfies the common support for the evaluation policy $\pi_e$ if
        $\pi_e(a|x) > 0 \implies \pi_b(a|x) > 0$ for all $x \in \calX$ and $a \in \calA$.
\end{assumption}
Assumption \ref{common_support} often holds in practice as long as the behavior policy assigns a non-zero probability to the action whose probability to be selected in the evaluation policy is non-zero. For the variance of IPS \cite{saito2022off}, we can decompose it into three terms as follows by using the law of total variance.
\begin{align*}
    n \mV_{\calD}\left[ \hat{V}_{\text{IPS}}(\pi_e; \calD) \right] ={}& \mE_{p(x)\pi_b(a|x)}[w(x, a)^2 \sigma(x, a)^2] \\
    +& \mV_{p(x)}\left[ \mE_{\pi_b(a|x)}[w(x, a) q(x, a)] \right] \\
    +& \mE_{p(x)}\left[ \mV_{\pi_b(a|x)}[w(x, a) q(x, a)] \right]
\end{align*}
where $\sigma(x, a) := \mV_{p(r|x, a)}[r]$ is the variance of the reward. Though the unbiasedness of IPS is preferable, the importance weight $w(x, a)$ has a wide range, resulting in a significant variance by the first and third terms when the cardinality of the action space is large. 

\subsection{Doubly Robust}
DR \cite{dudik2014doubly} combines the preferable properties of DM and IPS, defined as follows.
    \begin{align*}
        &\hat{V}_{\text{DR}}(\pi_e; \calD, \hat{q}) \\
        ={}& \frac{1}{n} \sum_{i=1}^n\left\{ \mE_{\pi_e(a|x_i)}[\hat{q}(x_i, a)] + w(x_i, a_i) \left( r_i - \hat{q}(x_i, a_i) \right) \right\}
    \end{align*}
 DR guarantees unbiasedness under either Assumption \ref{unbias_q_x_a} or \ref{common_support}. We can obtain the variance of DR by replacing $q(x, a)$ with $\Delta_{q, \hat{q}}(x, a)$ in the third terms of IPS variance where $\Delta_{q, \hat{q}}(x, a) := q(x, a) - \hat{q}(x, a)$ is the prediction error of the expected reward. Even though DR alleviates the variance of IPS, DR still incurs significant variance with large action spaces due to the first and third terms of DR variance.

\subsection{Marginalized Inverse Propensity Scoring}
To tackle the problem of the significant bias of DM and variance of IPS and DR under large action spaces, MIPS \cite{saito2022off} was proposed. Instead of using the importance weight $w(x, a)$ used in IPS, MIPS uses the marginal importance weight $w(x, e)$ where $e \in \calE \subset \mR^{d_e}$ is the embedding of the action. For instance, if action $a$ is the movie we recommend, the embedding $e$ can be the genre, directors, and actors which categorize the movie. If the embedding characterizes the film well, we can reduce the cardinality of embedding space $|\calE|$. Therefore, using the embedding for the marginal importance weight improves the variance of the MIPS. To use action embedding, we define the new data-generating process and value function as follows.

We sample the action embedding $e$ from an unknown distribution $e \sim p(e|x, a)$ given $x$ and $a$, and the reward from an unknown distribution $r \sim p(r|x, a, e)$ given $x$,  $a$, and $e$. Thus, the logged data $\calD$ is 
\begin{align*}
    \calD &= \{(x_i, a_i, e_i, r_i)\}_{i=1}^n \\
    &\sim \prod_{i =1}^n p(x_i) \pi_b (a_i|x_i) p(e_i|x_i, a_i) p(r_i|x_i, a_i, e_i).
\end{align*}
Following the new data-generating process, we define the value function $V(\pi)$ as follows.
\begin{align*}
    V(\pi) 
    &:= \mE_{p(x) \pi(a|x) p(e|x, a) p(r|x, a, e)}[r] \\
    &= \mE_{p(x) \pi(a|x) p(e|x, a)}[q(x, a, e)] \\
    &= \mE_{p(x) \pi(a|x)}[q(x, a)]
\end{align*}
where $q(x, a, e) := \mE_{p(r|x, a, e)}[r|x, a, e]$ is the expected reward function given $x, a$, and $e$ and $q(x, a) := \mE_{p(e|x, a)}[q(x, a, e)]$.

Having the new data-generating process and the definition of the value function,  MIPS \cite{saito2022off} is defined as follows.
\begin{align*}
    \mips := \frac{1}{n} \sum_{i = 1}^n w(x_i, e_i) r_i
\end{align*}
where $w(x, e) := p(e|x, \pi_e)/p(e|x, \pi_b)$ is the marginal importance weight and $p(e|x, \pi) := \sum_{a \in \calA} \pi(a|x) p(e|x, a)$ is the marginal distribution of $e$ given the context vector $x$ and policy $\pi$. MIPS is unbiased under two assumptions as follows.
\begin{assumption}[No Direct Effect of Action on Reward]
    \label{no_direct_effect}
    Given context $x$ and action embedding $e$, action $a$ and reward $r$ are independent $(a \indep r | x, e)$
\end{assumption}
No direct effect assumption means that the action embedding $e$ fully mediates the effect of action $a$ on the reward $r$.

\begin{assumption}[Common Embedding Support]
    \label{common_embed_support}
    We say that the data-generating process satisfies the common embedding support if $p(e|x, \pi_e) > 0 \implies p(e|x, \pi_b) > 0$ for all $x \in \calX$ and $e \in \calE$.
\end{assumption}
Common embedding support is a weaker assumption than the common support necessary for the IPS's unbiasedness. Even with the considerable variance reduction of MIPS against IPS, the unbiasedness of MIPS is not necessarily guaranteed in practice as Assumption \ref{no_direct_effect} usually does not hold due to the difficulty finding the perfect embedding of the action.

\section{Proposed estimator}
\begin{figure*}
    \centering
    \includegraphics[width=1\textwidth]{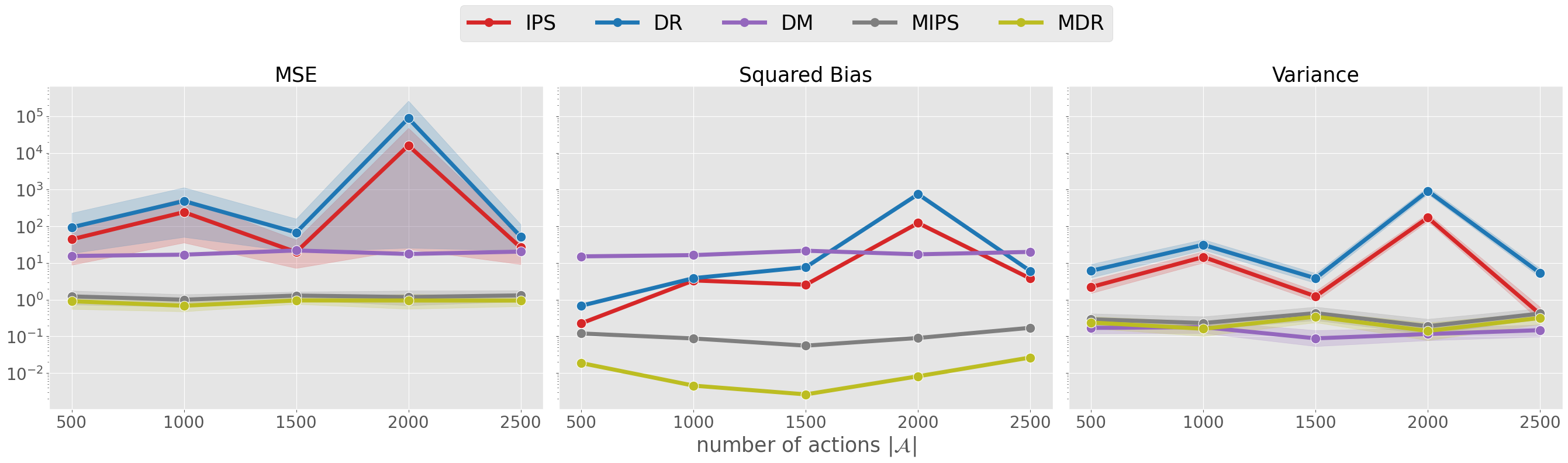}
    \caption{MSE (left), Bias (center), and variance (right) of DM, IPS, DR, MIPS, and MDR(ours) when we change the cardinality of the action space}
    \label{mdr}
\end{figure*}
In this section, we combine MIPS and DR to construct a more accurate estimator to overcome the shortcomings of DM, IPS, DR, and MIPS. 
\subsection{Marginalized Doubly Robust}
To preserve the variance reduction of MIPS against IPS while having the double robustness of DR, we propose the novel estimator called \textit{\textbf{M}arginalized \textbf{D}oubly \textbf{R}obust (\textbf{MDR})}  as follows.
\begin{align*}
    &\mdr \\
    ={}& \meanN \left\{ \mE_{\pi_e(a|x_i)}[\hat{q}(x_i, a)] + w(x_i, e_i) \left( r_i - \hat{q}(x_i, a_i, e_i) \right) \right\}
\end{align*}
The first term of MDR is the baseline estimator, which uses the regression model $\hat{q}$ to incorporate the parametric approach. The second term incorporates the importance weight $w(x_i, e_i)$ on the action embedding space to weight the residuals of the estimated expected reward function, having the doubly robust structure. 

Intuitively, MDR is supposed to estimate the value function $V(\pi_e)$ of the evaluation policy $\pi_e$ more accurately than the existing estimators. Compared to IPS, MDR uses importance weight $w(x, e)$ in the action embedding space $\calE$ whose cardinality is much smaller than the action space $\calA$ where the importance weight of IPS $w(x, a)$ is defined. Thus, MDR is supposed to have a much lower variance than IPS. In comparison with MIPS, MDR is supposed to be unbiased under the situation where MIPS is not because MDR incorporated the regression model $q(x, a, e)$ to capture the direct effect of action $a$ on the reward $r$ that MIPS ignored. We theoretically analyze the statistical properties of MDR in the following subsection.
\subsection{Theoretical Analysis}
MDR has a doubly robust property under either Assumptions \ref{no_direct_effect} and \ref{common_embed_support} or the following assumption about the precision of the prediction of the expected reward function $q(x, a, e)$.
\begin{assumption}[Perfect Estimation of Expected Reward given $x, a$, and $e$]
    \label{unbias_q_x_a_e} 
    We say regression model $\hat{q}(x, a, e)$ perfectly estimates the expected reward function $q$ if $\hat{q}(x, a, e) = q(x, a, e)$ for all $x \in \calX, a \in \calA$, and $e \in \calE$
\end{assumption}

\begin{proposition}[Unbiasedness of MDR]
    \label{mdr_unbias}
    MDR is unbiased under either Assumptions \ref{no_direct_effect} and \ref{common_embed_support}, or Assumption \ref{unbias_q_x_a_e}. See Appendix \ref{proof_unbiasedness_mdr} for the proof.
\end{proposition}

Proposition \ref{mdr_unbias} shows that MDR is more likely to be unbiased than MIPS as MDR requires weaker assumptions than MIPS due to its doubly robust structure. However, we cannot guarantee that either assumption holds, so we derive the bias of MDR when the assumptions required for unbiasedness are not satisfied. 

\begin{proposition}[Bias of MDR]
    \label{prop.bias_MDR}
   If Assumptions \ref{common_embed_support} is true, but Assumptions \ref{no_direct_effect} and \ref{unbias_q_x_a_e} are violated, then MDR has the following bias. 
   \begin{align*}
       &\bias (\mdr) \\
       ={}& \mE_{p(x) p(e|x, \pi_b)} \Bigg[ \sum_{a<b} \pi_b(a|x, e) \pi_b(b|x, e) \\
       & \times \big( \Delta_{q, \hat{q}}(x, a, e) -  \Delta_{q, \hat{q}}(x, b, e) \big) \times \left( w(x, b) - w(x, a) \right) \Bigg]
   \end{align*}
   See Appendix \ref{proof_bias_mdr} for the proof.
\end{proposition}
Compared to the bias of MIPS, MDR has the following bias reduction.
\begin{align*}
    &\bias (\mips) - \bias (\mdr) \\ 
    ={}& \mE_{p(x) p(e|x, \pi_b)} \Bigg[ \sum_{a<b} \pi_b(a|x, e) \pi_b(b|x, e) \\
    & \times \big( \hat{q}(x, a, e) - \hat{q}(x, b, e) \big) \times \left( w(x, b) - w(x, a) \right) \Bigg]
\end{align*}
In addition to the bias, we derived the variance of MDR. We compare the variance of MDR and MIPS and show that MDR has the lower variance than MIPS as follows.
\begin{proposition}[Variance Reduction of MDR against MIPS]
    \label{variance_mdr}
    Under Assumptions \ref{common_support}, \ref{no_direct_effect}, and \ref{common_embed_support}, the difference between the variances of MIPS and MDR is 
    \begin{align*}
        &n\left( \mV_{\calD}\left[ \mips \right] - \mV_{\calD}\left[ \mdr \right] \right) \\
        ={}& \mE_{p(x)} \Big[ \mV_{\pi_b(a|x) p(e|x, a)} \left[ w(x, e) q(x, a, e) \right] \\
        &- \mV_{\pi_b(a|x) p(e|x, a)} \left[ w(x, e) \Delta_{q, \hat{q}}(x, a, e) \right]  \Big]
    \end{align*}
    where $\Delta_{q, \hat{q}}(x, a, e) := q(x, a, e) - \hat{q}(x, a, e)$ is the estimation error given $x, a, e$.
    See Appendix \ref{proof_of_variance_reduction} for the proof.
\end{proposition}
Proposition \ref{variance_mdr} shows that the variance of MIPS does not depend on the estimated expected reward $\hat{q}(x, a, e)$ whereas MDR can reduce its variance due to the use of the estimated expected reward function. Moreover, the more precise the estimator of expected reward given context, action, and action embedding is, the lower the variance of MDR becomes. Thus, as long as the estimator of the expected reward $\hat{q}(x, a, e)$ is not too far from the true value $q(x, a, e)$, MDR reduces the variance of MIPS:
\begin{align*}
     \mV_{\calD}\left[ \mips \right] > \mV_{\calD}\left[ \mdr \right] .
\end{align*}

\section{Simulation Study}
\begin{figure*}
    \centering
    \includegraphics[width=1\textwidth]{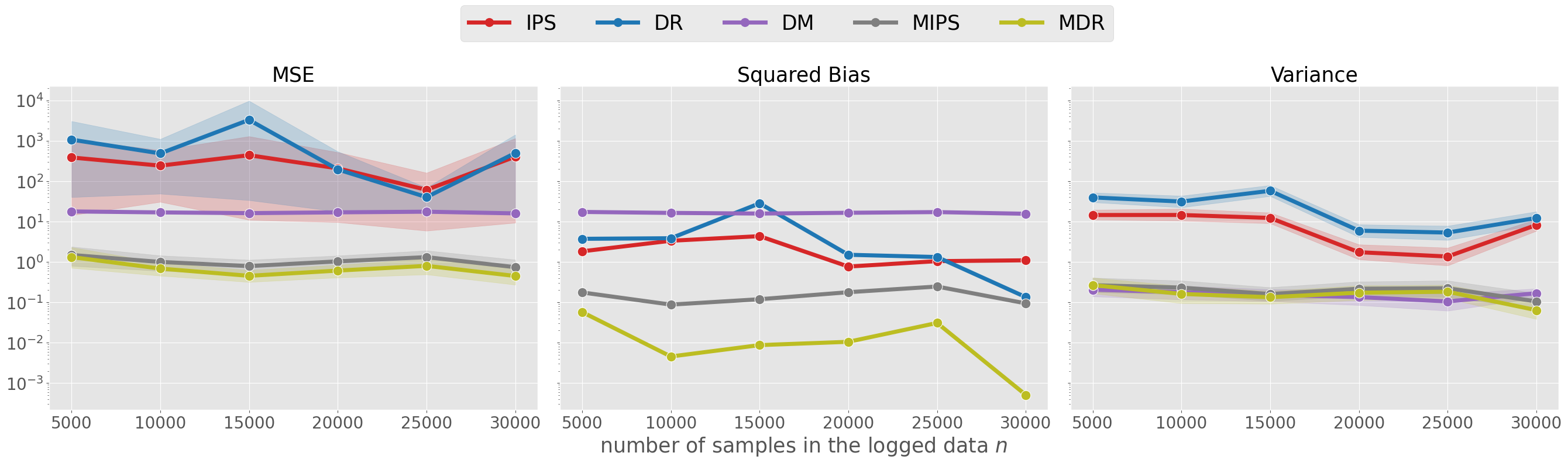}
    \caption{MSE (left), Bias (center), and variance (right) of DM, IPS, DR, MIPS, and MDR(ours) when we change the number of samples in the logged data}
    \label{fig_n_rounds}
\end{figure*}
We conducted the synthetic data experiment by implementing MDR by using the Open Bandit Pipeline \cite{saito2020open}. The Python code for the simulation can be found in the \href{https://github.com/tatsu432/DR-estimator-OPE-large-action}{https://github.com/tatsu432/DR-estimator-OPE-large-action}. The simulation environment is mostly the same as MIPS \cite{saito2022off}. 

\subsection{Synthetic Data}
In the synthetic data, we define the data-generating process as follows. The contextual vector $x$ is drawn i.i.d from the 10-dimensional standard normal distribution. Given action $a$, the embedding of the action is drawn i.i.d from the distribution 
\begin{align*}
    p(e|x, a) = \prod_{k \in [d_e]} \frac{\exp(\alpha_{a, e_k})}{\sum_{e' \in \calE_k} \exp(\alpha_{a, e'_k})}
\end{align*}
where $\alpha_{a, e_k}$ is a set of parameters drawn from the standard normal distribution $\calN(0, 1)$ and the cardinality of the action embedding space is 10; $\calE_k = [10]$. The behavior policy $\pi_b$ is then defined as 
\begin{align*}
    \pi_b(a|x) = \frac{\exp(\beta \cdot q(x, a))}{\sum_{a' \in \calA} \exp(\beta \cdot q(x, a'))}
\end{align*}
where $\beta$ is the parameter that handles the optimality of the behavior policy and $q(x, a) := \mE_{p(e|x, a)}[q(x, e)]$. The expected reward function $q(x, e)$ given context $x$ and action embedding $e$ is defined as 
\begin{align*}
    q(x, e) = \sum_{k \in [d_e]} \eta_k \cdot \left(x^\top M x_{e_k}+\theta_x^\top x + \theta_e^\top x_{e_k}\right)
\end{align*}
where $M, \theta_x$, and $\theta_e$ are parameters whose elements we sample from the uniform distribution whose range is $[-1, 1]$ and $\eta_k$ represents the importance of the $k$-th dimension of the action embedding sampled from Dirichlet distribution such that $\sum_{k \in [d_e]} \eta_k = 1$. Then the evaluation policy $\pi_e$ whose value function we want to estimate is defined as
\begin{align*}
    \pi_e(a|x) := (1-\epsilon) \cdot \ind \left\{ a = \argmax_{a' \in \calA} q(x, a') \right\} + \frac{\epsilon}{|\calA|}
\end{align*}
where $\epsilon \in [0, 1]$ represents the quality of the evaluation policy $\pi_e$. If we set it near zero, then the evaluation policy is near optimal, and we set it to $\epsilon = 0.05$. 

\subsection{Results}
\begin{figure*}
    \centering
    \includegraphics[width=1\textwidth]{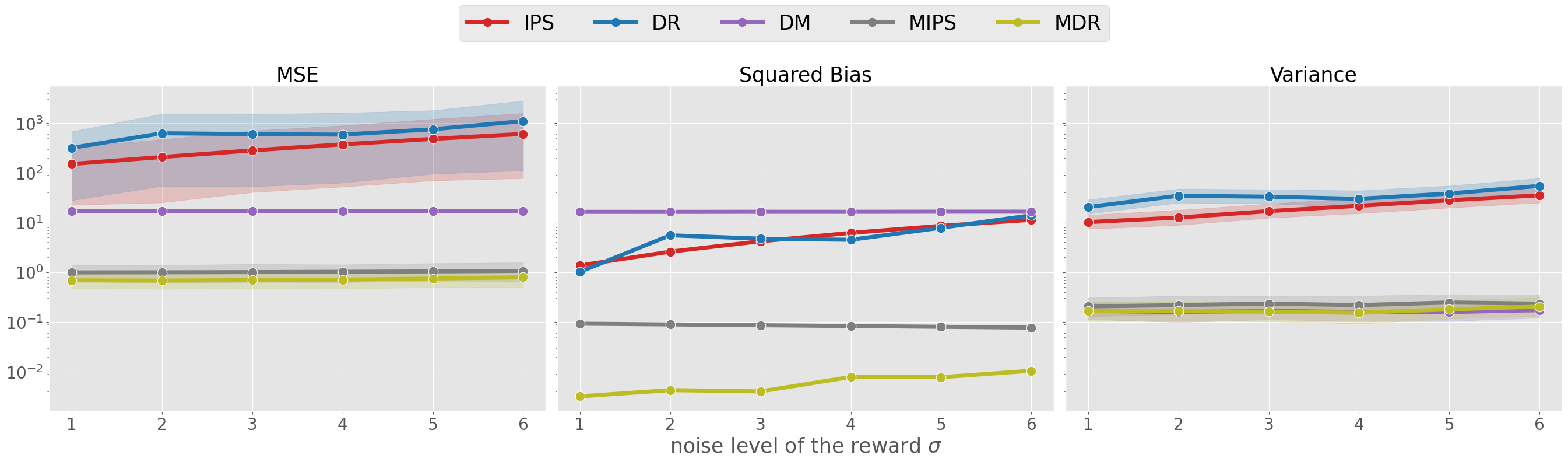}
    \caption{MSE (left), Bias (center), and variance (right) of DM, IPS, DR, MIPS, and MDR(ours) when we change the reward noise level}
    \label{fig_reward_std}
\end{figure*}
Figure 1 shows the MSE, bias, and variance of DM, IPS, DR, MIPS, and MDR. As the number of action spaces $|\calA|$ gets large, DM, IPS, and DR have high MSE. For DM, this is primarily because of the significant bias, whereas for IPS and DR, we can attribute this to the substantial variance caused by the wide range of the importance weight. MIPS and MDR overcome this problem by using the marginalized importance weight. Furthermore, MDR has a lower bias than MIPS due to the doubly robust property of MDR. Moreover, the variance of MDR is reduced against IPS and DR as well as MIPS. 

Figure 2 demonstrates the MSE, bias, and variance of DM, IPS, DR, MIPS, and MDR when we vary the number of sample data $n$. As the number of logged data increases, the precision of the estimator of the expected reward $\hat{q}(x, a, e)$ improves, which leads to the decrease of the term $(\Delta_{q, \hat{q}}(x, a, e) - \Delta_{q, \hat{q}}(x, b, e))$ of the bias of MDR where $a, b \in \calA$. The result of the experiment support this theory as the bias of MDR decreases as the number of sample data increases. Furhermore, the precision of the estimator of the expected reward contributes to the decrease of the variance of MDR by reducing the term $\Delta_{q, \hat{q}}(x, a, e)$ in the variance reduction formula in Proposition \ref{variance_mdr}. Empirical result corroborate the theory of variance reduction as the variance of MDR compared to MIPS decrease slightly with increasing number of the logged data $n$.

Figure 3 empirically demonstrate the precision of MDR compared to the existing estimators when the noise level of the reward $\sigma$ changes. If the reward is less noisy, the quality of the estimator of the expected reward $\hat{q}(x, a, e)$ becomes more precise estimator for the true expected reward $q(x, a, e$. Thus, the low noise level of the reward contirbutes to the lower bias of MDR. The empirical result support this theory as the bias of MDR increase as the reward becomes noise. 

\section{Conclusion and Future Work}
We studied OPE with large action spaces and proposed the MDR, which has the doubly robust property to be unbiased under weaker assumptions than MIPS and has a significantly lower variance than IPS and DR. The simulation study demonstrates that MDR outperforms other existing estimators, including MIPS.

Our study gives rise to several interesting future directions. It is crucial to find the better action embedding $e$ to have the standard embedding support, which is one of the assumptions for the unbiasedness of MDR, so in the future, it would be interesting to find the algorithm to construct the better embedding. Moreover, other estimators might be doubly robust under different assumptions. Thus, it would be intriguing to compare the candidates of MDR by simulation study or combine them to construct the triply robust estimator.

\section*{Acknowledgements} 
We thank Yuta Saito at Cornell University Department of Computer Science for his support and guidance.


\bibliographystyle{icml2023}
\bibliography{ref}

\newpage
\appendix
\onecolumn

\section{Proof of the unbiasedness of MDR}
\label{proof_unbiasedness_mdr}
\begin{proof}
First, when Assumptions \ref{no_direct_effect} and \ref{common_embed_support} are satisfied, we have 
\begin{align*}
    & \mE_{\calD}\left[ \mdr \right] \\ 
    ={}& \mE_{\calD}\left[ \mips + \meanN \big\{ \mE_{\pi_e(a|x_i)}[\hat{q}(x_i, a)] - w(x_i, e_i) \hat{q}(x_i, a_i, e_i)\big\} \right] \quad \because \text{definition of MDR} \\
    ={}& V(\pi_e) + \mE_{\calD}\bigg[ \meanN \big\{ \mE_{\pi_e(a|x_i)}[\hat{q}(x_i, a)] - w(x_i, e_i) \hat{q}(x_i, a_i, e_i)\big\} \bigg] \quad \because \text{unbiasedness of MIPS} \\
    ={}& V(\pi_e) + \mE_{p(x)\pi_b(a|x) p(e|x, a) p(r|x, a, e) }
    \bigg[  \mE_{\pi_e(a'|x)}[\hat{q}(x, a')] - w(x, e) \hat{q}(x, a, e) \bigg]  \quad \because \text{i.i.d. assumption} \\
    ={}& V(\pi_e) + \mE_{p(x)\pi_e(a'|x) }
    \left[ \hat{q}(x, a') \right] - \mE_{p(x)\pi_b(a|x) p(e|x, a)}
    \left[ w(x, e) \hat{q}(x, a, e) \right] \\
    ={}& V(\pi_e) + \mE_{p(x)\pi_e(a'|x) }
    \left[ \hat{q}(x, a') \right] - \mE_{p(x)}
    \left[\sumA \pi_b(a|x) \sumE p(e|x, a) \frac{p(e| x, \pi_e)}{p(e|x, \pi_b)} \hat{q}(x, e) \right] \quad \because \text{Assumption \ref{no_direct_effect}} \\
    ={}& V(\pi_e) + \mE_{p(x)\pi_e(a'|x) }
    \left[ \hat{q}(x, a') \right] - \mE_{p(x)}
    \left[ \sumE \frac{p(e| x, \pi_e)}{p(e|x, \pi_b)} \hat{q}(x, e) \sumA \pi_b(a|x) p(e|x, a) \right] \quad \because \text{order of sum} \\
    ={}& V(\pi_e) + \mE_{p(x)\pi_e(a'|x) }
    \left[ \hat{q}(x, a') \right] - \mE_{p(x)}
    \left[\sumE \frac{p(e| x, \pi_e)}{p(e|x, \pi_b)} \hat{q}(x, e) p(e|x, \pi_b) \right] \quad \because \text{definition of $p(e|x, \pi_b)$ } \\
    ={}& V(\pi_e) + \mE_{p(x)\pi_e(a'|x) }
    \left[ \hat{q}(x, a') \right] - \mE_{p(x)}
    \left[\sumE p(e| x, \pi_e) \hat{q}(x, e)  \right] \quad \because \text{cancel out $p(e|x, \pi_b)$ } \\
    ={}& V(\pi_e) + \mE_{p(x)\pi_e(a'|x) }
    \left[ \hat{q}(x, a') \right] - \mE_{p(x)} \left[\sumA \pi_e(a|x) \hat{q}(x, a)  \right] \\
    ={}& V(\pi_e) 
\end{align*}

Secondly, when the Assumption \ref{unbias_q_x_a_e} is satisfied, we have 
\begin{align*}
    &\mE_{\calD}\left[ \mdr \right] \\
    ={}& \mE_{\calD}\left[ \meanN \left\{ \mE_{\pi_e(a|x_i)}[\hat{q}(x_i, a)] + w(x_i, e_i) \left( r_i - \hat{q}(x_i, a_i, e_i) \right) \right\} \right] \quad \because \text{definition of MDR} \\
    ={}& \mE_{p(x)\pi_b(a|x) p(e|x, a) p(r|x, a, e)} \left[ \mE_{\pi_e(a'|x)}[\hat{q}(x, a')] + w(x, e) \left( r - \hat{q}(x, a, e) \right) \right] \quad \because \text{i.i.d. assumption} \\
    ={}& \mE_{p(x)\pi_b(a|x) p(e|x, a) p(r|x, a, e)} \left[ \mE_{\pi_e(a'|x)}[q(x, a')] + w(x, e) \left( r -q(x, a, e) \right) \right] \quad \because \text{Assumption \ref{unbias_q_x_a_e}} \\
    ={}& \mE_{p(x)\pi_b(a|x) p(e|x, a)} \left[ \mE_{\pi_e(a'|x)}[q(x, a')] + w(x, e) \left( q(x, a, e) - q(x, a, e) \right) \right]\\
    ={}& \mE_{p(x)\pi_b(a|x) p(e|x, a)} \left[ \mE_{\pi_e(a'|x)}[q(x, a')] \right]\\
    ={}& \mE_{p(x)} \left[ \mE_{\pi_e(a'|x)}[q(x, a')] \right]\\
    ={}& V(\pi_e) 
\end{align*}
\end{proof}

\section{Proof of the bias of MDR}
\label{proof_bias_mdr}
\begin{proof}
    If Assumption \ref{common_embed_support} holds, but Assumptions \ref{no_direct_effect} and \ref{unbias_q_x_a_e} are violated, we can show the bias of MDR as follows.
    \begin{align*}
        & \bias \left( \mdr \right) \\
        ={}& \mE_{\calD} \left[ \mdr \right] - V(\pi_e) \quad \because \text{definition of bias of MDR} \\
        ={}& \mE_{\calD} \left[ \mips + \meanN \big\{ \mE_{\pi_e(a|x_i)}[\hat{q}(x_i, a)] - w(x_i, e_i) \hat{q}(x_i, a_i, e_i)\big\} \right] - V(\pi_e) \quad \because \text{definition of MDR} \\
        ={}& \bias \left( \mips \right) + \mE_{\calD} \left[ \meanN \big\{ \mE_{\pi_e(a|x_i)}[\hat{q}(x_i, a)] - w(x_i, e_i) \hat{q}(x_i, a_i, e_i)\big\} \right] \quad \because \text{def. of bias of MIPS} \\
        ={}& \bias \left( \mips \right) + \mE_{p(x) \pi_b(a|x) p(e|x, a) p(r|x, a, e)} \left[\mE_{\pi_e(a'|x)}[\hat{q}(x, a')] - w(x, e) \hat{q}(x, a, e) \right] \quad \because \text{i.i.d. assumption} \\
        ={}& \bias \left( \mips \right) - \mE_{p(x)} \left[ \mE_{\pi_b(a|x) p(e|x, a)}[w(x, e) \hat{q}(x, a, e)] + \mE_{\pi_e(a|x) p(e|x, a)}[\hat{q}(x, a, e)]  \right] \\
        ={}& \bias \left( \mips \right) - \mE_{p(x)} \left[ \sumA \pi_b(a|x) \sumE p(e|x, a) w(x, e) \hat{q}(x, a, e) \right] \\
        & + \mE_{p(x)} \left[ \sumA \pi_e(a|x) \sumE p(e|x, a) \hat{q}(x, a, e) \right] \\
        ={}& \bias \left( \mips \right) - \mE_{p(x)} \left[ \sumA \pi_b(a|x) \sumE \frac{p(e|x, \pi_b) \pi_b(a|x, e)}{\pi_b(a|x)} w(x, e) \hat{q}(x, a, e) \right] \\
        & + \mE_{p(x)} \left[ \sumA \pi_e(a|x) \sumE \frac{p(e|x, \pi_b) \pi_b(a|x, e)}{\pi_b(a|x)} \hat{q}(x, a, e) \right] \quad \because \text{$p(e|x, \pi_b) = \frac{p(e|x, \pi_b) \pi_b(a|x, e)}{\pi_b(a|x)}$} \\
        ={}& \bias \left( \mips \right) - \mE_{p(x)} \left[ \sumE p(e|x, \pi_b) w(x, e) \sumA \pi_b(a|x, e) \hat{q}(x, a, e) \right] \\
        & +  \mE_{p(x)} \left[ \sumE p(e|x, \pi_b) \sumA w(x, a) \pi_b(a|x, e) \hat{q}(x, a, e) \right] \\
        ={}& \bias \left( \mips \right) - \mE_{p(x) p(e|x, \pi_b)} \left[ w(x, e) \sumA \pi_b(a|x, e) \hat{q}(x, a, e) \right] \\
        & +  \mE_{p(x) p(e|x, \pi_b)} \left[ \sumA w(x, a) \pi_b(a|x, e) \hat{q}(x, a, e) \right] \\
        ={}& \bias \left( \mips \right) - \mE_{p(x) p(e|x, \pi_b)} \left[ \sumA w(x, a) \pi_b(a|x,e) \sum_{b \in \calA} \pi_b(b|x, e) \hat{q}(x, b, e) \right] \\
        & +  \mE_{p(x) p(e|x, \pi_b)} \left[ \sumA w(x, a) \pi_b(a|x, e) \hat{q}(x, a, e) \right] \quad \because \text{$w(x, e) =  \sumA w(x, a) \pi_b(a|x,e)$} \\
        ={}& \bias \left( \mips \right) - \mE_{p(x) p(e|x, \pi_b)} \left[ \sumA w(x, a) \pi_b(a|x,e) \left( \left( \sum_{b \in \calA} \pi_b(b|x, e) \hat{q}(x, b, e) \right) - \hat{q}(x, a, e) \right) \right] 
    \end{align*}
    \begin{align*}
        ={}&  \mE_{p(x) p(e|x, \pi_b)} \Bigg[ \sum_{a<b} \pi_b(a|x, e) \pi_b(b|x, e)  \times \big( q(x, a, e) - q(x, b, e) \big) \times \left( w(x, b) - w(x, a) \right) \Bigg] \\
        &-  \mE_{p(x) p(e|x, \pi_b)} \Bigg[ \sum_{a<b} \pi_b(a|x, e) \pi_b(b|x, e)  \times \big( \hat{q}(x, a, e) - \hat{q}(x, b, e) \big) \times \left( w(x, b) - w(x, a) \right) \Bigg] \quad \because \text{bias of MIPS} \\
        ={}&  \mE_{p(x) p(e|x, \pi_b)} \Bigg[ \sum_{a<b} \pi_b(a|x, e) \pi_b(b|x, e)  \times \big( \Delta_{q, \hat{q}}(x, a, e) - \Delta_{q, \hat{q}}(x, b, e) \big) \times \left( w(x, b) - w(x, a) \right) \Bigg] \\
    \end{align*}
\end{proof}
where we use the Lemma B.1. by \citet{saito2022off} for the second last equality.

\newpage

\section{Proof of the variance reduction of MDR}
\label{proof_of_variance_reduction}
\begin{proof}
    Under Assumptions \ref{common_support}, \ref{no_direct_effect}, and \ref{common_embed_support} (i.e., the common support, no direct effect of action on reward, and the common embedding support), we can derive the variance reduction of MDR against MIPS as follows.
    \begin{align*}
        & n\left( \mV_{\calD}\left[ \mips \right] - \mV_{\calD}\left[ \mdr \right] \right) \\
        ={}& n\left( \mV_{\calD}\left[ \meanN w(x_i, e_i) r_i \right] - \mV_{\calD}\left[ \meanN \left\{ \mE_{\pi_e(a|x_i)}[\hat{q}(x_i, a)] + w(x_i, e_i) \left( r_i - \hat{q}(x_i, a_i, e_i) \right) \right\} \right] \right)  \quad \because \text{def. of MIPS and MDR}  \\
        ={}& \mV_{p(x) \pi_b(a|x) p(e|x, a) p(r|x, a, e)} \left[w(x, e) r \right] \\
        & -  \mV_{p(x) \pi_b(a|x) p(e|x, a) p(r|x, a, e)} \left[ \mE_{\pi_e(a'|x))} \left[ \hat{q}(x, a') \right] + w(x, e) (r - \hat{q}(x, a, e) \right] \quad \because \text{i.i.d. assumption} \\
        ={}& \mV_{p(x) \pi_b(a|x) p(e|x, a) p(r|x, a, e)} \left[w(x, e) r \right] \\
        & - \mV_{p(x) \pi_b(a|x) p(e|x, a)} \left[ \mE_{p(r|x, a, e)} \left[ \mE_{\pi_e(a'|x))} \left[ \hat{q}(x, a') \right] + w(x, e) (r - \hat{q}(x, a, e) \right] \right] \\
        & - \mE_{p(x) \pi_b(a|x) p(e|x, a)} \left[ \mV_{p(r|x, a, e)} \left[ \mE_{\pi_e(a'|x))} \left[ \hat{q}(x, a') \right] + w(x, e) (r - \hat{q}(x, a, e) \right] \right] \quad \because \text{total variance law} \\
        ={}& \mV_{p(x) \pi_b(a|x) p(e|x, a) p(r|x, a, e)} \left[w(x, e) r \right] - \mV_{p(x) \pi_b(a|x) p(e|x, a)} \left[ \mE_{\pi_e(a'|x))} \left[ \hat{q}(x, a') \right] + w(x, e) \Delta_{q, \hat{q}}(x, a, e) \right] \\
        & - \mE_{p(x) \pi_b(a|x) p(e|x, a)} \left[ \mV_{p(r|x, a, e)} \left[ w(x, e) r \right] \right] \\
        ={}& \mV_{p(x) \pi_b(a|x) p(e|x, a)} \left[ \mE_{p(r|x, a, e)} \left[ w(x, e) r \right] \right] - \mV_{p(x) \pi_b(a|x) p(e|x, a)} \left[ \mE_{\pi_e(a'|x))} \left[ \hat{q}(x, a') \right] + w(x, e) \Delta_{q, \hat{q}}(x, a, e) \right] \quad \because \text{total var. law} \\
        ={}& \mV_{p(x) \pi_b(a|x) p(e|x, a)} \left[ w(x, e) q(x, a, e) \right] - \mV_{p(x) \pi_b(a|x) p(e|x, a)} \left[ \mE_{\pi_e(a'|x))} \left[ \hat{q}(x, a') \right] + w(x, e) \Delta_{q, \hat{q}}(x, a, e) \right] \\
        ={}& \mV_{p(x)} \left[ \mE_{\pi_b(a|x) p(e|x, a)} \left[ w(x, e) q(x, a, e) \right] \right] + \mE_{p(x)} \left[ \mV_{\pi_b(a|x) p(e|x, a)} \left[ w(x, e) q(x, a, e) \right] \right] \\
        & - \mV_{p(x)} \left[ \mE_{\pi_b(a|x) p(e|x, a)} \left[ \mE_{\pi_e(a'|x))} \left[ \hat{q}(x, a') \right] + w(x, e) \Delta_{q, \hat{q}}(x, a, e) \right] \right] \\
        & - \mE_{p(x)} \left[ \mV_{\pi_b(a|x) p(e|x, a)} \left[\mE_{\pi_e(a'|x))} \left[ \hat{q}(x, a') \right] + w(x, e) \Delta_{q, \hat{q}}(x, a, e) \right] \right] \quad \because \text{total var. law} \\
        ={}& \mV_{p(x)} \left[ \mE_{\pi_b(a|x) p(e|x, a)} \left[ w(x, e) q(x, a, e) \right] \right]  - \mV_{p(x)} \left[ \mE_{\pi_b(a|x) p(e|x, a)} \left[ \mE_{\pi_e(a'|x))} \left[ \hat{q}(x, a') \right] + w(x, e) \Delta_{q, \hat{q}}(x, a, e) \right] \right] \\
        & + \mE_{p(x)} \left[ \mV_{\pi_b(a|x) p(e|x, a)} \left[ w(x, e) q(x, a, e) \right] - \mV_{\pi_b(a|x) p(e|x, a)} \left[ w(x, e) \Delta_{q, \hat{q}}(x, a, e) \right] \right] \\
        ={}& \mE_{p(x)} \left[ \mV_{\pi_b(a|x) p(e|x, a)} \left[ w(x, e) q(x, a, e) \right] - \mV_{\pi_b(a|x) p(e|x, a)} \left[ w(x, e) \Delta_{q, \hat{q}}(x, a, e) \right] \right] \\
        & + \mV_{p(x)} \left[ \sumA \sumE \pi_b(a|x) p(e|x, a) w(x, e) q(x, a, e) \right] \\
        & - \mV_{p(x)} \left[  \mE_{\pi_e(a'|x))} \left[ \hat{q}(x, a') \right] + \sumA \sumE \pi_b(a|x) p(e|x, a) w(x, e) \Delta_{q, \hat{q}}(x, a, e) \right] \\
        ={}& \mE_{p(x)} \left[ \mV_{\pi_b(a|x) p(e|x, a)} \left[ w(x, e) q(x, a, e) \right] - \mV_{\pi_b(a|x) p(e|x, a)} \left[ w(x, e) \Delta_{q, \hat{q}}(x, a, e) \right] \right] \\
        & + \mV_{p(x)} \left[ \sumA \sumE \pi_b(a|x) p(e|x, a) w(x, e) q(x, e) \right]  \\
        & - \mV_{p(x)} \left[  \mE_{\pi_e(a'|x))} \left[ \hat{q}(x, a') \right] + \sumA \sumE \pi_b(a|x) p(e|x, a) w(x, e) \Delta_{q, \hat{q}}(x, e) \right] \quad \because \text{Assumption \ref{no_direct_effect}} \\
        ={}& \mE_{p(x)} \left[ \mV_{\pi_b(a|x) p(e|x, a)} \left[ w(x, e) q(x, a, e) \right] - \mV_{\pi_b(a|x) p(e|x, a)} \left[ w(x, e) \Delta_{q, \hat{q}}(x, a, e) \right] \right] \\
        & + \mV_{p(x)} \left[ \sumE w(x, e) q(x, e) \sumA \pi_b(a|x) p(e|x, a) \right]  \\
        & - \mV_{p(x)} \left[  \mE_{\pi_e(a'|x))} \left[ \hat{q}(x, a') \right] + \sumE w(x, e) \Delta_{q, \hat{q}}(x, e) \sumA \pi_b(a|x) p(e|x, a) \right] \quad \because \text{change the order of sum} \\
    \end{align*}
    \begin{align*}
        ={}& \mE_{p(x)} \left[ \mV_{\pi_b(a|x) p(e|x, a)} \left[ w(x, e) q(x, a, e) \right] - \mV_{\pi_b(a|x) p(e|x, a)} \left[ w(x, e) \Delta_{q, \hat{q}}(x, a, e) \right] \right] + \mV_{p(x)} \left[ \sumE w(x, e) q(x, e) p(e|x, \pi_b) \right]  \\
        & - \mV_{p(x)} \left[  \mE_{\pi_e(a'|x))} \left[ \hat{q}(x, a') \right] + \sumE w(x, e) \Delta_{q, \hat{q}}(x, e) p(e|x, \pi_b) \right] \quad \because \text{definition of $p(e|x, \pi_b)$} \\
        ={}& \mE_{p(x)} \left[ \mV_{\pi_b(a|x) p(e|x, a)} \left[ w(x, e) q(x, a, e) \right] - \mV_{\pi_b(a|x) p(e|x, a)} \left[ w(x, e) \Delta_{q, \hat{q}}(x, a, e) \right] \right] + \mV_{p(x)} \left[ \sumE p(e|x, \pi_e) q(x, e) \right]  \\
        & - \mV_{p(x)} \left[  \mE_{\pi_e(a'|x))} \left[ \hat{q}(x, a') \right] + \sumE p(e|x, \pi_e) \Delta_{q, \hat{q}}(x, e) \right] \quad \because \text{cancel out $p(e|x, \pi_b)$} \\
        ={}& \mE_{p(x)} \left[ \mV_{\pi_b(a|x) p(e|x, a)} \left[ w(x, e) q(x, a, e) \right] - \mV_{\pi_b(a|x) p(e|x, a)} \left[ w(x, e) \Delta_{q, \hat{q}}(x, a, e) \right] \right] \\
        & + \mV_{p(x)} \left[ \sumE p(e|x, \pi_e) q(x, e) \right] - \mV_{p(x)} \left[  \mE_{\pi_e(a'|x))} \left[ \hat{q}(x, a') \right] + \sumE p(e|x, \pi_e) \Delta_{q, \hat{q}}(x, e) \right] \\
        ={}& \mE_{p(x)} \left[ \mV_{\pi_b(a|x) p(e|x, a)} \left[ w(x, e) q(x, a, e) \right] - \mV_{\pi_b(a|x) p(e|x, a)} \left[ w(x, e) \Delta_{q, \hat{q}}(x, a, e) \right] \right] \\
        & + \mV_{p(x)} \left[ \sumE p(e|x, \pi_e) q(x, a, e) \right] - \mV_{p(x)} \left[  \mE_{\pi_e(a'|x))} \left[ \hat{q}(x, a') \right] + \sumE p(e|x, \pi_e) \Delta_{q, \hat{q}}(x, a, e) \right] \quad \because \text{Assumption \ref{no_direct_effect}} \\
        ={}& \mE_{p(x)} \left[ \mV_{\pi_b(a|x) p(e|x, a)} \left[ w(x, e) q(x, a, e) \right] - \mV_{\pi_b(a|x) p(e|x, a)} \left[ w(x, e) \Delta_{q, \hat{q}}(x, a, e) \right] \right] \\
        & + \mV_{p(x)} \left[ \sumA \pi_e(a|x) \sumE p(e|x, a) q(x, a, e) \right] - \mV_{p(x)} \left[  \mE_{\pi_e(a'|x))} \left[ \hat{q}(x, a') \right] + \sumA \pi_e(a|x) \sumE p(e|x, a) \Delta_{q, \hat{q}}(x, a, e) \right] \\
        ={}& \mE_{p(x)} \left[ \mV_{\pi_b(a|x) p(e|x, a)} \left[ w(x, e) q(x, a, e) \right] - \mV_{\pi_b(a|x) p(e|x, a)} \left[ w(x, e) \Delta_{q, \hat{q}}(x, a, e) \right] \right] \\
        & + \mV_{p(x)} \left[ \mE_{\pi_e(a|x)} \left[ q(x, a) \right] \right] - \mV_{p(x)} \left[  \mE_{\pi_e(a'|x))} \left[ \hat{q}(x, a') \right] + \mE_{\pi_e(a|x)} \left[ \Delta_{q, \hat{q}}(x, a) \right] \right] \\
        ={}& \mE_{p(x)} \left[ \mV_{\pi_b(a|x) p(e|x, a)} \left[ w(x, e) q(x, a, e) \right] - \mV_{\pi_b(a|x) p(e|x, a)} \left[ w(x, e) \Delta_{q, \hat{q}}(x, a, e) \right] \right] \\
        & + \mV_{p(x)} \left[ \mE_{\pi_e(a|x)} \left[ q(x, a) \right] \right] - \mV_{p(x)} \left[ \mE_{\pi_e(a|x)} \left[ q(x, a) \right] \right] \\
        ={}& \mE_{p(x)} \left[ \mV_{\pi_b(a|x) p(e|x, a)} \left[ w(x, e) q(x, a, e) \right] - \mV_{\pi_b(a|x) p(e|x, a)} \left[ w(x, e) \Delta_{q, \hat{q}}(x, a, e) \right] \right]
    \end{align*}
\end{proof}

\section{Additional Results on Synthetic Bandit Data}

\begin{figure*}
    \centering
    \includegraphics[width=1\textwidth]{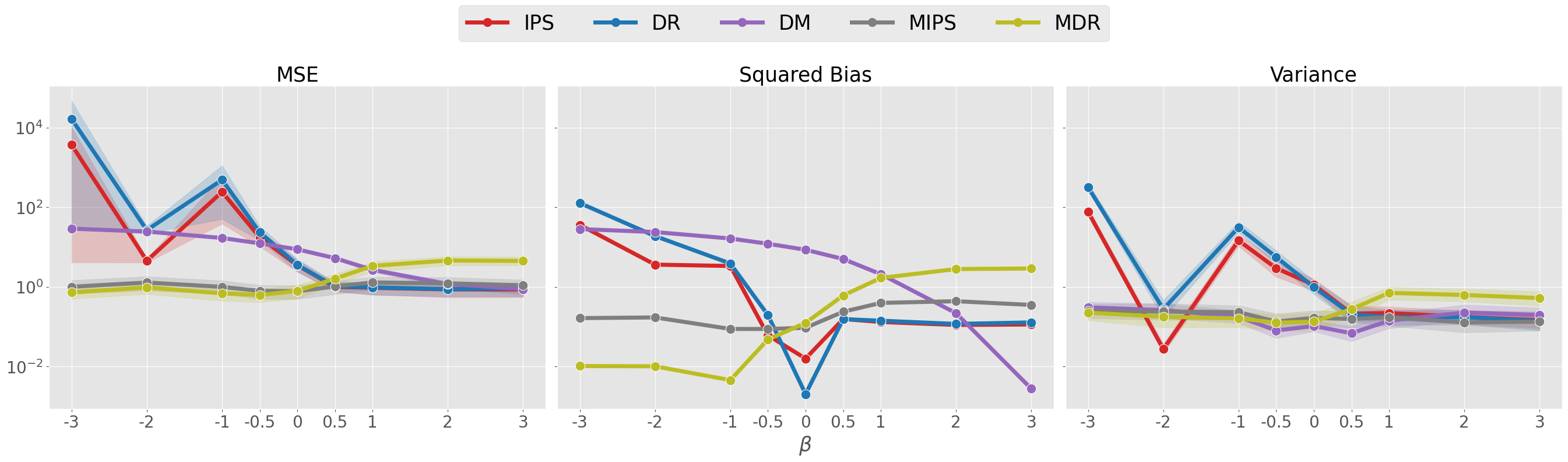}
    \caption{MSE (left), Bias (center), and variance (right) of DM, IPS, DR, MIPS, and MDR(ours) when we change the logging policy (beta)}
    \label{fig_beta}
\end{figure*}

\begin{figure*}
    \centering
    \includegraphics[width=1\textwidth]{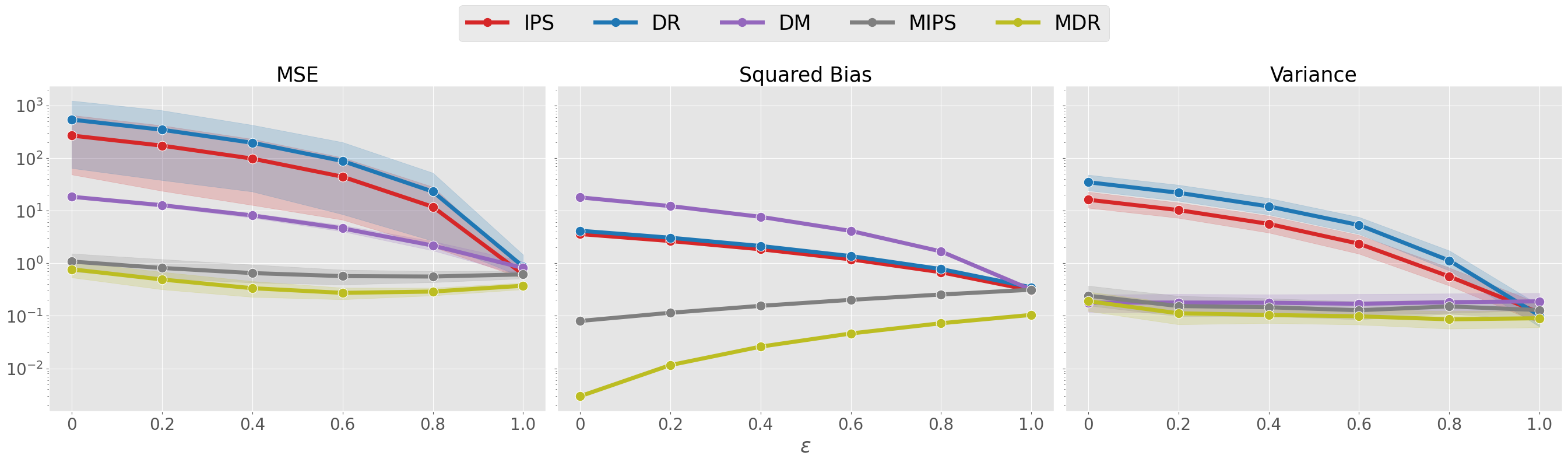}
    \caption{MSE (left), Bias (center), and variance (right) of DM, IPS, DR, MIPS, and MDR(ours) when we change the evaluation policy (epsilon)}
    \label{fig_epsilon}
\end{figure*}

\begin{figure*}
    \centering
    \includegraphics[width=1\textwidth]{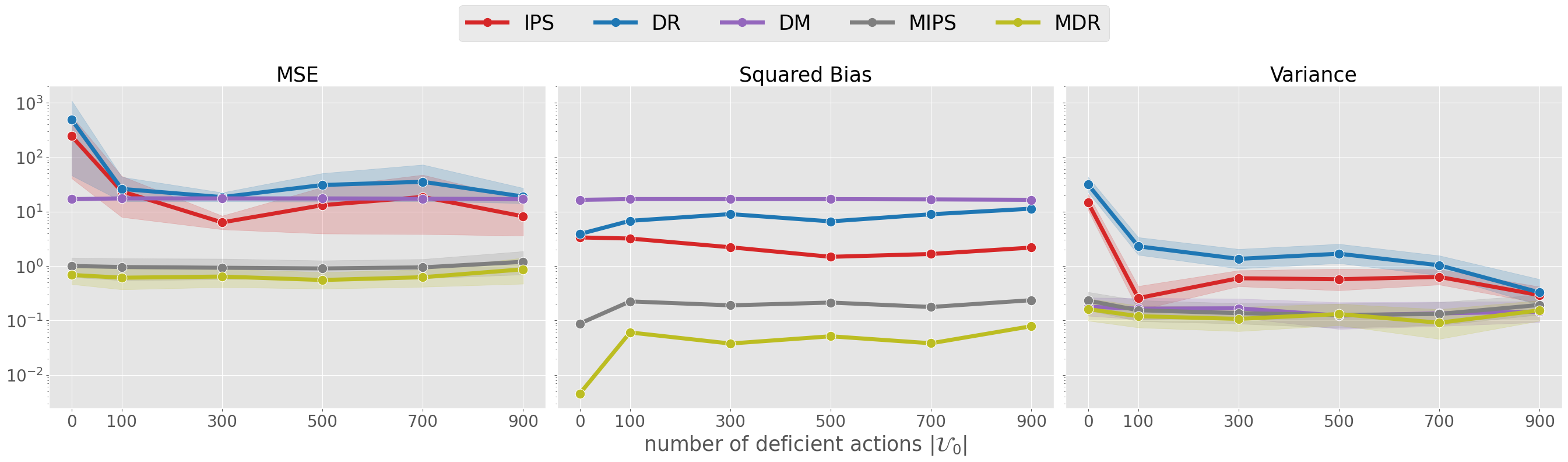}
    \caption{MSE (left), Bias (center), and variance (right) of DM, IPS, DR, MIPS, and MDR(ours) when we change the number of deficient actions}
    \label{fig_n_deficient_actions}
\end{figure*}

\begin{figure*}
    \centering
    \includegraphics[width=1\textwidth]{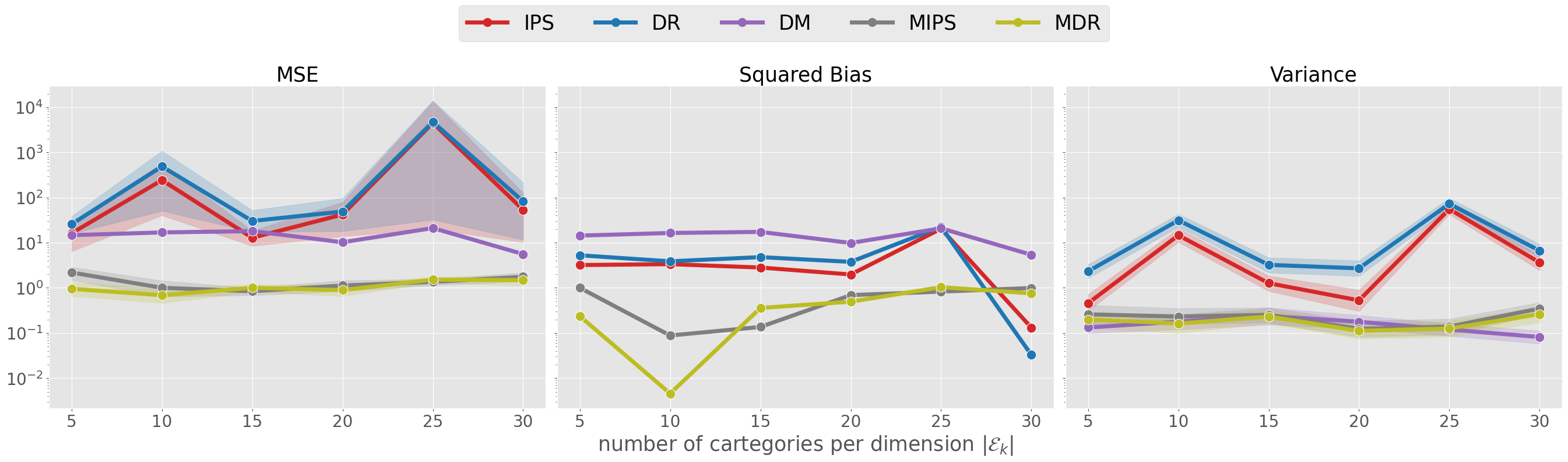}
    \caption{MSE (left), Bias (center), and variance (right) of DM, IPS, DR, MIPS, and MDR(ours) when we change the number of categories per dimension}
    \label{fig_n_cat_per_dim}
\end{figure*}

\begin{figure*}
    \centering
    \includegraphics[width=1\textwidth]{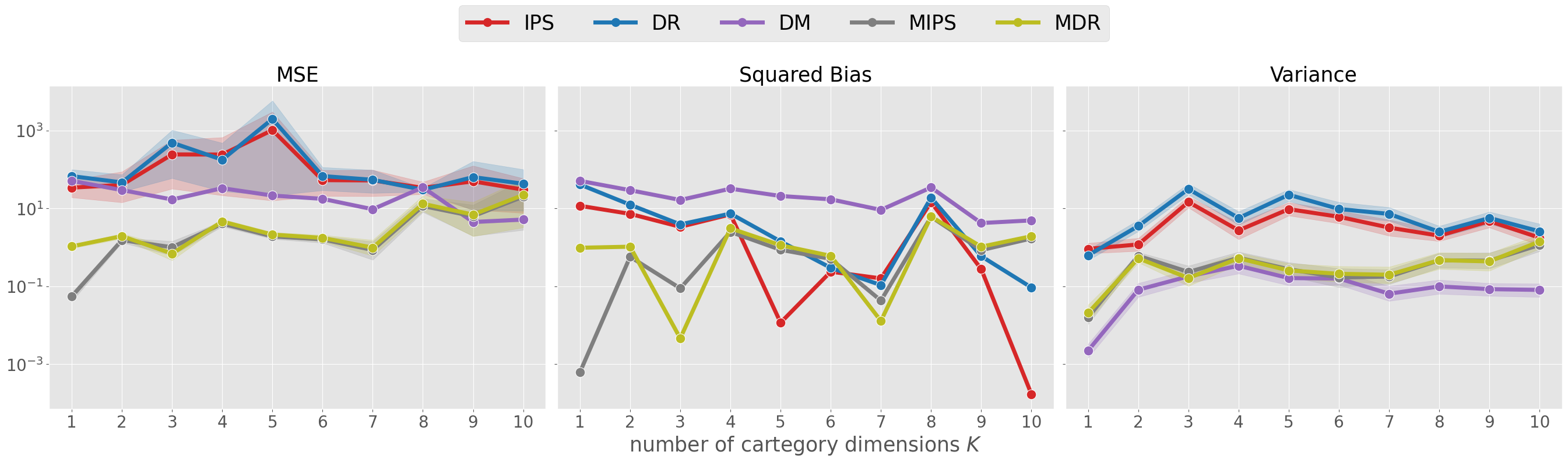}
    \caption{MSE (left), Bias (center), and variance (right) of DM, IPS, DR, MIPS, and MDR(ours) when we change the number of the category dimension}
    \label{fig_n_cat_dim}
\end{figure*}


\end{document}